\documentclass[jair,twoside,11pt,theapa]{article}
\usepackage{jair, theapa, rawfonts}

\jairheading{x}{xxxx}{xx-xx}{xx/xx}{xx/xx}

\ShortHeadings{Community-based 3-SAT Formulas with a Predefined Solution}
{Hu, Luo, \& Wang}
\firstpageno{1}

\usepackage{algorithm}
\usepackage{multirow}
\usepackage{array}
\usepackage{algpseudocode}
\usepackage{amsmath}
\usepackage{booktabs}
\usepackage{graphicx}
\usepackage{float}
\usepackage{amssymb}
\usepackage{boldline}
\usepackage{url}

\newcommand{\myTable}{Table. }
\newcommand{\myFigure}{Fig. }
\newcommand{\myAlgorithm}{Alg. }
\newcommand*\wrapletters[1]{\wr@pletters#1\@nil}
\def\wr@pletters#1#2\@nil{#1\allowbreak\if&#2&\else\wr@pletters#2\@nil\fi}

\begin{document}
	
	\title{Community-based 3-SAT Formulas \\ with a Predefined Solution}
	
	\author{\name Yamin Hu \email huym@mail.ustc.edu.cn \\
		\name Wenjian Luo \email wjluo@ustc.edu.cn \\
		\name Junteng Wang \email wjt2013@mail.ustc.edu.cn \\
		\addr Anhui Province Key Laboratory of Software Engineering in Computing and Communication \\
		School of Computer Science and Technology\\
		University of Science and Technology of China, China
	}
	
	\maketitle
	
	\begin{abstract}
		It is crucial to generate crafted SAT formulas with predefined solutions for the testing and development of SAT solvers since many SAT formulas from real-world applications have solutions.
		Although some generating algorithms have been proposed to generate SAT formulas with predefined solutions, community structures of SAT formulas are not considered in these algorithms.
		Consequently, we propose a 3-SAT formula generating algorithm that not only guarantees the existence of a predefined solution, but also simultaneously considers community structures and clause distributions.
		The proposed 3-SAT formula generating algorithm controls the quality of community structures through controlling
		(1) the number of clauses whose variables have a common community, which we call intra-community clauses,
		and 
		(2) the number of variables that only belong to one community, which we call intra-community variables.
		For a SAT formula, more intra-community clauses and intra-community variables, higher quality of community structures.
		To study the combined effect of community structures and clause distributions on the hardness of SAT formulas, we measure solving runtimes of two solvers, gluHack (a leading CDCL solver) and CPSparrow (a leading SLS solver), on the generated SAT formulas under different groups of parameter settings. Through extensive experiments, we obtain some noteworthy observations on the SAT formulas generated by the proposed algorithm: 
		(1) The community structure has little or no effects on the hardness of SAT formulas with regard to CPSparrow but a strong effect with regard to gluHack. 
		(2) Only when the proportion of true literals in a SAT formula in terms of the predefined solution is 0.5, SAT formulas are hard-to-solve with regard to gluHack; when this proportion is below 0.5, SAT formulas are hard-to-solve with regard to CPSparrow.
		(3) When the ratio of the number of clauses to that of variables is around 4.25, the SAT formulas are hard-to-solve with regard to both gluHack and CPSparrow.
		
	\end{abstract}
	
	\section{Introduction}
	
	The Boolean satisfiability problem (sometimes called SAT), i.e., determining whether a given Boolean formula is satisfiable or not, is the first proven NP-complete problem \cite{cook1971complexity}. 
	The study of SAT problem has attracted attentions from many computer scientists, because the SAT problem has extensive range of practical applications, such as hardware design and verification \cite{gupta2006sat}.
	
	A SAT formula is a Boolean formula over a set of Boolean variables (denoted as $ V $). 
	In SAT formulas, a literal is a variable such as $ v $, called positive literal, or the negation of a variable such as $ \stackrel{-}{v} $, called negative literal. 
	The polarity of a literal is the sign of the corresponding variable; 
	that is to say, the polarity of a positive literal is positive, while the polarity of a negative literal is negative. 
	If a literal is true in terms of the corresponding variable assignment, then it is called true literal; otherwise, false literal. 
	In 3-SAT formulas in conjunctive normal form (CNF), a clause is a disjunction of 3 literals, i.e., $ C = v_{i}(or \stackrel{-}{v_{i}}) \vee v_{j} (or \stackrel{-}{v_{j}}) \vee v_{k} (or \stackrel{-}{v_{k}}) $, where $ 1 \le i, j, k \le n $ ($ n $ is the number of variables in the 3-SAT formula),
	and a formula is a conjunction of clauses, i.e., $ \varphi = C_{1} \wedge C_{2} \wedge \cdots \wedge C_{m} $ ($ m $ is the number of clauses in the 3-SAT formula). 
	The formulas generated by the proposed generating algorithm in this paper are 3-SAT formulas in conjunctive normal form.
	A predefined solution is the assignments to all variables in $ V $ that satisfy all clauses in $ \varphi $, where every clause has at least one true literal. 
	
	In recent years, many solvers have been proposed to solve SAT formulas, and the components in these solvers have been becoming more and more complicated \cite{balyo2017sat}. 
	The currently popular and successful SAT solvers include CDCL (Conflict-Driven Clause Learning) solvers and SLS (Stochastic Local Search) solvers. 
	Each type of solvers has both strengths and weaknesses.
	Inspired by DPLL (Davis-Putnam-Logemann-Loveland) backtracker \cite{davis1962machine}, CDCL solvers were proposed. 
	Through learning new clauses by conflict analyses and backtracking nonchronologically, CDCL solvers could find solutions or prove no solution.
	That is to say, CDCL solvers are complete.
	CDCL solvers are good at solving industrial formulas, so that it have greatly promoted the applications of SAT problems \cite{giraldez2016generating}.
	The popular CDCL solvers include gluHack \cite{heule2018proceedings}, MiniSAT \cite{een2003extensible}, ZChaff \cite{mahajan2004zchaff2004}, etc.
	In addition, look-ahead based SAT solvers are also based on the DPLL backtracker \cite{heule2009look}.
	Different from CDCL solvers, in order to find a solution, SLS solvers simply flip a variable to make more clauses satisfiable (i.e., greedy strategy),
	or randomly flip a variable to avoid being stuck in local optimums (i.e., random strategy).
	SLS solvers usually perform well on random SAT formulas and use fewer memory than CDCL solvers \cite{balint2009novel}.
	The popular SLS solvers include WalkSat \cite{selman1993local}, CPSparrow \cite{belov2014proceedings}, etc.
	
	The performance of newly proposed solvers is measured on many groups of SAT formulas, which are usually called benchmarks \cite{balyo2017sat,audemard2016extreme}.
	Furthermore, this process needs a substantial number and a variety of benchmarks \cite{hoos2000satlib}.
	These benchmarks were divided into application formulas (also known as real-world or industrial formulas), and random crafted formulas. 
	In this paper, we focus on the generation of crafted formulas, which could greatly increase the types of crafted SAT formulas.
	
	Many generating algorithms of SAT formulas have been proposed \cite{giraldez2016generating,achlioptas2000generating,blsi2008random,burg2012creating}. Some were proposed to generate SAT formulas with some property, such as the high-quality community structure (see Subsection \ref{subsec-com} for details), and the power law distribution in the numbers of occurrences of variables in SAT formulas \cite{ansotegui2009towards}. 
	Note that the generating algorithms of SAT formulas are essential and extremely important for the testing of development of SAT solvers.	
	However, these generating algorithms have some drawbacks, such as they cannot guarantee the existence of solutions in the resulting SAT formulas. 
	It is worth mentioning that SAT formulas with solutions are more useful for the testing of incomplete solvers \cite{achlioptas2000generating}. 
	The reason is that, for a SAT formula with solutions, when an incomplete solver does not find any solution in bounded time, we could ensure that the performance of the solver is low, instead of containing no solution in the given SAT formula.
	In the following paragraph, we introduce some generating algorithms that can generate SAT formulas with predefined solutions.
	
	The generating algorithms of SAT formulas with predefined solutions mainly include: the 1-hidden algorithm \cite{achlioptas2005hiding}, the 2-hidden algorithm \cite{achlioptas2005hiding}, the $ q $-hidden algorithm \cite{jia2005generating}, the $ p $-hidden algorithm \cite{liu2014p}, and the $ K $-hidden algorithm \cite{zhao2015fine,zhao2017experimental}.
	These algorithms generate clauses one by one.
	For a clause, these algorithms first select variables by simple random sampling without replacement from the set of all Boolean variables.
	Then, these algorithms assign polarities (positive or negative) to selected variables, which will construct a clause. 
	According to the number of true literals, clauses are divided into different types.
	According to some probability model, these algorithms generate some type of clause by assigning polarities, which is the reason for the name of clause distribution.
	The algorithms ensure the existence of predefined solutions by filtering out unsatisfiable clauses in terms of the predefined solution.
	The difference between these algorithms is the approaches used to assigning polarities to variables of a clause, which are explained below.
	The 1-hidden algorithm \cite{achlioptas2005hiding} assigns every variable of a clause to positive or negative polarity with equal probability; if the resulting clause is unsatisfiable, just remove and regenerate it.
	However, the polarities of literals of the resulting SAT formula are biased, so that solvers might obtain a correct assignment of a variable by simply counting the numbers of positive and negative literals corresponding to the variable with high probability \cite{achlioptas2005hiding}.
	Consequently, the resulting formulas are usually easy to solve.
	In order to remove the bias in the 1-hidden algorithm, the 2-hidden algorithm \cite{achlioptas2005hiding} was proposed, which simultaneously filters out clauses in which all literals are unsatisfiable or satisfiable.
	Later on, the $ q $-hidden algorithm \cite{jia2005generating} was proposed to generate hard-to-solve 3-SAT formulas with regard to SLS solvers by hiding solutions deceptively.
	The $ q $-hidden algorithm use one parameter to control clause distributions.
	Following the $ q $-hidden algorithm, the $ p $-hidden algorithm \cite{liu2014p} was proposed, which is an extension of the $ q $-hidden algorithm.
	The $ p $-hidden algorithm use two parameters to control clause distributions, and it has wider parameter space than the $ q $-hidden algorithm.
	Thus the $ p $-hidden algorithm can generate harder-to-solve formulas with regard to SLS solvers than the $ q $-hidden algorithm.
	Besides, the $ K $-hidden algorithm \cite{zhao2015fine,zhao2017experimental} was proposed to generate $ K $-SAT formulas, which have fine-grained control for clause distributions.
	However, these algorithms do not consider community structures of SAT formulas.
	
	In this paper, we propose a novel 3-SAT formula generating algorithm. 
	Through guaranteeing that clauses in resulting formulas are all satisfiable in terms of a predefined solution, the proposed algorithm can ensure the existence of the predefined solution. 
	Also, the proposed algorithm can control the numbers of different types of clauses (i.e., clause distributions) of SAT formulas.
	Besides, the proposed algorithm has finer control of community structures. That is to say, it can control community structures by controlling the number of intra-community clauses and the number of intra-community variables at the same time.
	For clarity, the main contributions of this paper are given as follows. 
	
	\begin{itemize}
		\item[1)] We propose a novel 3-SAT formula generating algorithm with a predefined solution, which considers both community structures and clause distributions. The proposed algorithm controls community structures through controlling the numbers of both intra-community clauses and intra-community variables.
		\item[2)] Through extensive experiments, we study the hardness of the generated 3-SAT formulas with regard to both gluHack and CPSparrow under different groups of parameter settings, and obtain some noteworthy observations.
	\end{itemize}
	
	The remainder of this paper is organized as follows. 
	In Section \ref{sec-relatedwork}, we introduce two kinds of SAT formula generating algorithms, which are related to the proposed 3-SAT formula generating algorithm in this paper. 
	In Section \ref{sec-theproposedalgorithm}, we describe the proposed generating algorithm in detail.
	In Section \ref{sec-experiments}, through experiments, we test and analyze the hardness of SAT formulas generated by our generating algorithm with regard to gluHack and CPSparrow under different groups of parameter settings. 
	In Section \ref{sec-discussions}, we present some discussions related to the proposed generating algorithm. 
	In Section \ref{sec-conclusions}, we conclude this paper, and present our future work.
	
	\section{Related work}
	\label{sec-relatedwork}
	In this section, we present two types of SAT formula generating algorithms strongly related to the proposed 3-SAT formula generating algorithm in this paper. 
	The former considers community structures, and the latter considers clause distributions.
	
	\subsection{Generating algorithms considering community structures}
	\label{subsec-com}
	Industrial SAT formulas are considered to have distinct natures with random uniform-k-SAT formulas, such as community structures \cite{giraldez2016generating}.
	The quality of community structures is usually measured by modularity; higher modularity means higher quality of community structures. 
	With high probability, the modularity of random uniform-k-SAT formulas is low, while the modularity of industrial SAT formulas is high \cite{ansotegui2012community}. The community structure of industrial SAT formulas is correlated with the solving runtimes of CDCL SAT solvers \cite{newsham2014impact,mull2016hardness,zulkoski2018effect}.
	A typical algorithm that could generate SAT formulas with controllable quality of community structures is called Community Attachment \cite{giraldez2016generating}.
	Here, we explain Community Attachment in detail, which could generate SAT formulas with community structures of a specified modularity.
	Community Attachment interprets SAT formulas as Variable Incidence Graph (VIG) \cite{ansotegui2012community}. In VIG, nodes are variables, and there is an edge between two nodes if they appear in one clause.
	In Community Attachment, modularity is calculated by \cite{giraldez2016generating}
	\begin{equation}
	\label{e-q}
	\begin{split}
	Q(G, P) =  \sum_{C_{i} \in P} \left( \dfrac{\sum_{x,y \in C_{i}} w(x,y)}{\sum_{x,y \in V} w (x,y)}- \left( \dfrac{\sum_{x \in C_{i}}deg(x)}{\sum_{x \in V}deg(x)} \right) ^{2} \right),
	\end{split}
	\end{equation}
	where $ G $ is the VIG of a SAT formula, $ P $ is a partition of nodes of the graph $ G $, $ C_{i} $ is the $i$-th community in the partition $ P $, $ V $ is the set of nodes in the graph $ G $, $ w(x,y) $ is the weight between nodes $ x $ and $ y $, and $ deg(x) $ is the degree of the node $ x $. For 3-SAT formulas, an edge corresponds the weight of $ \frac{1}{3} $, so that the weight between two nodes is the number of edges between these two nodes times $ \frac{1}{3} $.
	
	In Community Attachment, when generating a clause, with probability $ p = Q + \frac{1}{c} $ ($ Q $ is the parameter of Community Attachment that denotes the value of a preset modularity), variables are selected from a randomly selected community;
	with probability $ 1 - p $, $ k $ variables are selected from $ k $ randomly selected communities respectively.
	It has been proven that the modularity of resulting SAT formulas is around the preset modularity \cite{giraldez2016generating}, which validates the correctness of the above procedure.
	In the proposed 3-SAT formula generating algorithms, we will adopt a similar procedure to control the number of intra-community clauses.
	
	However, in Community Attachment, the polarities of variables are set to positive or negative with equal probability, that is to say, this algorithm does not guarantee the existence of a predefined solution and does not consider clause distributions.
	In the proposed algorithm in this paper, besides community structures, we guarantee the existence of a predefined solution and consider clause distributions.
	
	\subsection{Generating algorithms considering clause distributions}
	
	This type of generating algorithms are usually used to generate $ k $-SAT formulas with predefined solutions. 
	According to the number of true literals in a clause in terms of the predefined solution, clauses are divided into $ k + 1 $ types (denoted as Type 0, Type 1, Type 2, $ \cdots $, Type $ k $).
	In clauses of Type $ k $, there are $ k $ true literals.
	These algorithms ensure the existence of predefined solutions by filtering out clauses of Type 0.
	In these algorithms, when generating a clause, first, variables are randomly selected from the set of all Boolean variables, then the polarities of the selected variables, which determines the type of the resulting clause, are set according to some clause distribution.
	The $ q $-hidden algorithm \cite{jia2005generating} uses a parameter (i.e., $ q $) to control the clause distribution. When the $ q $-hidden algorithm is used to generate 3-SAT formula, with probability  $ p_{1} = \dfrac{3q}{(1+q)^3-1} $, a clause of Type 1 is generated; with probability  $ p_{2} = \dfrac{3q^{2}}{(1+q)^3-1} $, a clause of Type 2 is generated; with probability  $ p_{3} = 1-\dfrac{3q}{(1+q)^3-1}-\dfrac{3q^{2}}{(1+q)^3-1} $, a clause of Type 3 is generated.
	The $ p $-hidden algorithm \cite{liu2014p} is used to generate 3-SAT formulas, which uses two parameters (i.e., $ {p}_{1} $ and $ {p}_{2} $) to control clause distributions;
	when generating a clause, with probability  $ {p}_{1} $, a clause of Type 1 is generated;
	with probability  $ {p}_{2} $, a clause of Type 2 is generated;
	with probability  (1 $ -  $ $ {p}_{1} $ $ - $ $ {p}_{2} $) , a clause of Type 3 is generated.
	As can be seen above, the $ q $-hidden algorithm is a special case of the $ p $-hidden algorithm, and these two algorithms ensure the existence of a predefined solution by not generating clauses of Type 0.
	Note that SAT formulas generated by the $ p $-hidden algorithm could be harder-to-solve than that generated by the $ q $-hidden algorithm with regard to SLS solvers, because SLS solvers are more likely to be misguided to a region without the predefined solution on SAT formulas generated by the $ p $-hidden algorithm than that generated by the $ q $-hidden algorithm \cite{liu2014p}.
	In addition, by controlling the numbers of $ k $ types of clauses through $ k $ probability parameters \{$ p_{1}, p_{2}, ..., p_{k} $\}, the $ K $-hidden algorithm \cite{zhao2015fine,zhao2017experimental} could generate $ k $-SAT formulas with a predefined solution.
	
	However, in these algorithms, community structures of SAT formulas are not considered.
	In the proposed algorithm in this paper, besides clause distributions, community structures of SAT formulas are also considered.
	
	\section{The proposed algorithm}
	\label{sec-theproposedalgorithm}
	In this section, we first introduce some symbols that are needed to describe the proposed 3-SAT formula generating algorithm. 
	The proposed 3-SAT formula generating algorithm simultaneously takes community structures and clause distributions into consideration. 
	Through filtering out unsatisfiable clauses (part of clause distribution), the proposed generating algorithm can guarantee the existence of a predefined solution. 
	The process is divided into two steps: (1) partitioning variables into communities (described in Subsection \ref{subsection-partition}), and (2) generating clauses (described in Subsection \ref{subsection-clause}), including selecting variables from one or three communities and assigning polarities for these variables. 
	
	The existing SAT formula generating algorithms considering community structures only consider disjoint communities \cite{giraldez2016generating}. 
	In order to simulate more real SAT applications, we explicitly consider overlapping communities \cite{lu2018extending} by simultaneously controlling the numbers of intra-community clauses and intra-community variables in the SAT formulas.
	In the proposed generating algorithm, clauses are divided into two types: intra-community clauses and inter-community clauses. The former means clauses whose variables have a common community, while the latter means clauses whose variables belongs to two or three communities.
	We use a method similar with Community Attachment \cite{giraldez2016generating} to control the number of intra-community clauses (controlled by the parameter $ p $).
	Meanwhile, variables are divided into intra-community variables or inter-community variables; 
	in this procedure, we use the parameter $ \alpha $ to control the number of intra-community variables. 
	Thus, for a generated 3-SAT formula, the expectation of the number of intra-community clauses is $ n*r*p $ ($ r $ is the ratio of the number of clauses to the number of variables), the expectation of the number of inter-community clauses is $ n*r*(1-p) $, the expectation of the number of intra-community variables is $ n*\alpha $; and the expectation of the number of inter-community variables is $ n*(1-\alpha) $. Note that, in the proposed algorithm, a variable belongs to at most two communities, which could be extended into multiple communities.
	
	\subsection{Symbols}
	The symbols used in this paper are listed as follows.
	\begin{itemize}
		\item $ \varphi $: a SAT formula;
		\item $ V $: a set of Boolean variables;
		\item $ v $: a variable;
		\item $ v_{i} $: the $ i $-th variable in $ V $;
		\item $ Cla $: a clause;
		\item $ Cla_{i} $: the $ i $-th clause in $ \varphi $;
		\item $ C $: a community;
		\item $ C_{i} $: the $ i $-th community;
		\item $ n $: the number of variables;
		\item $ m $: the number of clauses;
		\item $ p $: the ratio of intra-community clauses to all clauses in a SAT formula;
		\item $ \alpha $: the ratio of intra-community variables to all variables in a SAT formula;
		\item $ \beta $: the ratio of true literals to all literals in a SAT formula;
		\item $ r $: the ratio of clauses to variables, i.e., $ m/n $;
		\item $ c $: the number of communities;
		\item $ s $: a predefined solution of $ \varphi $ on $ V $.
	\end{itemize}
	
	\subsection{Partitioning variables into communities}
	\label{subsection-partition}
	In the proposed algorithm, we first partition $ n $ variables into $ c $ communities with the properties below.
	
	\begin{itemize}
		\item[(1)] Every variable has the equal probability to appear in some community.
		\item[(2)] Every variable has the equal probability to become a intra-community variable or a inter-community variable.
		\item[(3)] The intra-community variables are evenly distributed among the communities.
		\item[(4)] The inter-community variables are also evenly distributed among the communities.
	\end{itemize}
	
	Thus, we could generate SAT formulas with randomness (Property (1) and (2)) and balance (Property (3) and (4)) well. 
	To obtain the above properties, we propose an algorithm, called PartitionCommunity, as shown in \myAlgorithm \ref{alg-partition}. 
	The inputs are the number of variables $ n $, the number of communities $ c $, and the proportion of intra-community variables $ \alpha $. 
	The outputs of PartitionCommunity are $ cToVsMap $ and $ vToCsMap $, which describe a community partition.
	The two data structures of $ cToVsMap $ and $ vToCsMap $ provide convenience for subsequent operations.
	$ cToVsMap $ is the mapping from a community to a set of variables; 
	conversely, $ vToCsMap $ is the mapping from a variable to a set of communities. 
	For example, $ cToVsMap[C_{1}] = \{v_{1}, v_{2}\} $ means that the community $ C_{1} $ consists of the variables $ v_{1} $ and $ v_{2} $; 
	conversely, $ vToCsMap[v_{1}] = \{C_{1}, C_{2}\} $ means that the variable $ v_{1} $ belongs to the communities $ C_{1} $ and $ C_{2} $, so that $ v_{1} $ is a inter-community variable.
	
	\begin{algorithm}
		\caption{\textit{PartitionCommunity}: Partition variables into communities}
		\label{alg-partition}
		\textbf{Input:} $ n $, $ c $, $ \alpha $ \\
		\textbf{Output:} $ cToVsMap $, $ vToCsMap $
		\begin{algorithmic}[1]
			\State $ V \gets \{v_1, v_2, ..., v_n\} $
			\For{$ i \gets 1 $ \textbf{to} $ c $}
			\State $ cToVsMap[C_i] \gets \Call{Sample}{V,n/c} $
			\For{$ v $ \textbf{in} $ cToVsMap[C_i] $}
			\State $ vToCsMap[v] \gets \{C_i\} $
			\EndFor
			\State $ V \gets V - cToVsMap[C_i] $
			\EndFor
			\For{$ i \gets 1 $ \textbf{to} $ c $}
			\State $ interCVSet \gets \Call{Sample}{cToVsMap[C_i], n/c*(1-\alpha)} $	
			\For{$ v $ \textbf{in} $ interCVSet $}
			\State $ otherCSet \gets \{C_1,C_2,...,C_c\}-\{C_i\} $
			\State $ otherC \gets \Call{Sample}{otherCSet,1} $
			\State $ cToVsMap[otherC] \gets cToVsMap[otherC] \cup \{v\} $	
			\State $ vToCsMap[v] \gets vToCsMap[v] \cup \{otherC\} $		
			\EndFor
			\EndFor
			\State \Return $ cToVsMap $, $ vToCsMap $
		\end{algorithmic}
	\end{algorithm}
	
	In the pseudo-code of \myAlgorithm \ref{alg-partition}, $ V $ is the set of $ n $ variables.
	$ \Call{Sample}{set, n} $ returns a set of $ n $ elements randomly selected from $ set $ if $ n > 1 $, or one element if $ n = 1 $. 
	For simplicity of description, in the pseudo-code, we assume that $ n $ is divisible by $ c $, and $ n/c*\alpha $ is an integer.
	At Lines 2--8, all $ n $ variables are evenly distributed into $ c $ communities. At this point, all variables are intra-community variables. 
	At the $ i $-th iteration of $ c $ iterations, randomly select $ n/c $ variables into the $ i $-th community $ C_{i} $.
	At Lines 9--10, convert some intra-community variables to inter-community variables.
	In the $ i $-th iteration of $ c $ iteration, first select $ n/c*(1-\alpha) $ variables from the $ i $-th community as inter-community variables (the remaining variables are intra-community variables); 
	then, for every inter-community variable, randomly select a community except for the community it belongs to and assign the current inter-community variable to the selected community.
	
	\subsection{Generating clauses}
	\label{subsection-clause}
	Based on the community partition generated by \myAlgorithm \ref{alg-partition}, clauses are generated one by one. Every clause is generated through two steps described as follows.
	
	\begin{itemize}
		\item[Step 1:] Select three variables from one or three communities, which results in intra-community clause or inter-community clause. The parameter $ p $ controls the proportion of intra-community clauses.
		\item[Step 2:] Determine the polarities of every selected variables, i.e., positive or negative. This step ensures that there is a predefined solution in the resulting 3-SAT formula through filtering out clauses of Type 0, and controls its clause distribution through the parameters $ p_{1} $ and $ p_{2} $.
	\end{itemize}
	
	As can be seen, our algorithm considers community structures (Step 1), clause distributions (Step 2), and could generate 3-SAT formulas with a predefined solution (Step 2). 
	Thus, the proposed algorithm could be used to study the combined effect of community structures and clause distributions on the hardness of SAT formulas.
	
	The pseudo-code of the proposed 3-SAT formula generating algorithm is shown in \myAlgorithm \ref{alg-generator}.
	There are three groups of parameters: (1) The parameters in the first group are relevant to the community structure, including the proportion of intra-community clauses $ p $, the proportion of intra-community variables $ \alpha $, and the number of communities $ c $; (2) The parameters in the second group are used to control clause distributions, including $ {p}_{1} $ (the proportion of clauses of Type 1) and $ {p}_{2} $ (the proportion of clauses of Type 2); (3) Other parameters consist of the predefined solution $ s $, the ratio of the number of clauses to that of variables $ r = m/n $, and the number of variables $ n $. The output is the resulting 3-SAT formula $ \varphi $.
	\begin{algorithm}
		\caption{The proposed 3-SAT formula generating algorithm}
		\label{alg-generator}
		\textbf{Input:}  $ p $, $ \alpha $, $ c $, $ {p}_{1} $, $ {p}_{2} $, $ s $, $ r $, $ n $\\
		\textbf{Output:} 3-SAT formula $ \varphi $
		\begin{algorithmic}[1]
			\State $ \varphi $ $ \gets $ empty formula
			\State $ cToVsMap,vToCsMap \gets \Call{PartitionCommunity}{n,c,\alpha} $
			\For{$ i \gets 1 $ \textbf{to} $ round(r*n) $}
			\If{ \Call{Rand}{\,}\ $ \le\ p $ }
			\State $ vSet \gets \Call{SelectOne}{cToVsMap,vToCsMap} $
			\Else
			\State $ vSet \gets \Call{SelectThree}{cToVsMap,vToCsMap} $
			\EndIf
			\State $ randNum \gets $ \Call{Rand}{\,}
			\If{ $randNum \le\ $$ {p}_{1} $ }
			\State $ Cla \gets \Call{SetPolarity}{s,vSet,1}$ 
			\ElsIf{$randNum \le\ $$ {p}_{1} + {p}_{2} $}
			\State $ Cla \gets \Call{SetPolarity}{s,vSet,2}$ 
			\Else
			\State $ Cla \gets \Call{SetPolarity}{s,vSet,3}$ 
			\EndIf
			\State $ \varphi \gets \varphi \bigwedge Cla $
			\EndFor
			\State \Return $ \varphi $
		\end{algorithmic}
	\end{algorithm}
	
	In this pseudo-code of \myAlgorithm \ref{alg-generator}, $ \Call{Rand}{\,} $ returns a random float number, which is drawn on the interval $ [0, 1) $; $ \Call{SetPolarity}{s,vSet,num} $ returns a clause, which is generated through the following three steps:
	
	\begin{itemize}
		\item[1)] $ num $ variables are selected from variable set $ vSet $ (containing 3 variables) by simple random sampling without replacement.
		\item[2)] The selected variables remain the same (leading to positive literals) if they are TRUE in the predefined solution, and become its negation (leading to negative literals) if they are FALSE in the predefined solution. The remaining variables in $ vSet $ remain the same (leading to positive literals) if they are FALSE in the predefined solution, and become its negation (leading to negative literals) if they are TRUE in the predefined solution.
		\item[3)] Disjunction of the resulting 3 literals is the clause to return.
	\end{itemize}

	When selecting variables from one or three communities (code at Line 5 and Line 7 in \myAlgorithm \ref{alg-generator}, respectively), our goal is to make every variable have equal degree in general, which could make the resulting SAT formulas hard-to-solve in worst cases. The pseudo-code of $ \Call{SelectOne}{cToVsMap, vToCsMap} $ is shown in \myAlgorithm \ref{alg-select1}, where $ \Call{Sample}{ $  $\{C_1,C_2,...,C_c\}, num} $ randomly selects $ num $ communities from the communities $ \{C_1,...,C_c\} $, and $ \Call{SampleDiff}{collection, $  $ num} $ randomly selects $ num $ different elements from $ collection $. The code at Line 1 selects one community from all communities as the target community. In order to achieve the goal (equal degree), we first initialize a empty variable list $ vList $. It is noted that two elements in $vList$ could be the same. 
	Then, for each variable in the target community, if it is an intra-community variable (i.e., the condition at Line 4 is satisfied), we add it to the variable list twice; otherwise (i.e., it is an inter-community variable), add once. The reason for doing so is that inter-community variables occur in two communities.
	Finally, randomly select three different variables from $ vList $, and return the set of the selected variables.
	
	\begin{algorithm}
		\caption{SelectOne: Select variables from one community}
		\label{alg-select1}
		\textbf{Input:} $ cToVsMap $, $ vToCsMap $ \\
		\textbf{Output:} $ vSet $
		\begin{algorithmic}[1]
			\State $ targetC \gets \Call{Sample}{\{C_1,C_2,...,C_c\},1} $
			\State $ vList \gets $ empty list
			\For{$ v $ \textbf{in} $ cToVsMap[targetC] $}
			\If{$ \Call{Size}{vToCsMap[v]} = 1$}
			\State $ vList.\Call{append}{v} $
			\State $ vList.\Call{append}{v} $
			\Else
			\State $ vList.\Call{append}{v} $
			\EndIf
			\EndFor
			\State $ vSet \gets \Call{SampleDiff}{vList,3} $
			\State \Return $ vSet $
		\end{algorithmic}
	\end{algorithm}
	
	The pseudo-code of $ \Call{SelectThree}{cToVsMap, vToCsMap} $ is shown in \myAlgorithm \ref{alg-select3}. The code at Line 1 selects three communities from all communities as the target communities. In order to achieve the goal (equal degree) above, we first initialize an empty variable list $ vList $. 
	Then, for each variable in the three target communities, if it is an intra-community variable (i.e., the condition at Line 5 is satisfied), we add it to $ vList $ twice; otherwise (i.e., it is an inter-community variable), add once. 
	Finally, randomly select three different variables from $ vList $ with a constraint that the three variables do not belong to the same one community (i.e., the condition at Line 14), and return the set of the selected variables.

	\begin{algorithm}
		\caption{SelectThree: Select variables from different communities}
		\label{alg-select3}
		\textbf{Input:} $ cToVsMap $, $ vToCsMap $ \\
		\textbf{Output:} $ vSet $
		\begin{algorithmic}[1]
			\State $ targetCSet \gets \Call{Sample}{\{C_1,C_2,...,C_c\},3} $
			\State $ vList \gets  $ empty list
			\For{$ targetC $ in $ targetCSet $}
			\For{$ v $ in $ cToVsMap[targetC] $}
			\If{$ \Call{Size}{vToCsMap[v]} = 1$}
			\State $ vList.\Call{append}{v} $
			\State $ vList.\Call{append}{v} $
			\Else
			
			\State $ vList.\Call{append}{v} $
			\EndIf
			\EndFor
			\EndFor
			\State $ vSet \gets \Call{SampleDiff}{vList,3} $
			\While{($ \exists c, vSet \subseteq cToVsMap[c] $)}
			\State $ vSet \gets \Call{SampleDiff}{vSet,3} $
			\EndWhile
			\State \Return $ vSet $
		\end{algorithmic}
	\end{algorithm}
	
	After introducing the above two algorithms, we describe the proposed 3-SAT formula generating algorithm, i.e., \myAlgorithm \ref{alg-generator}. The code at Lines 4--8 controls the quality of community structure of SAT formulas.
	With probability  $ p $, variables that are used to construct a clause are selected from the same community (see \myAlgorithm \ref{alg-select1}); with probability  $ 1-p $, variables are selected from three communities (see \myAlgorithm \ref{alg-select3}). After selecting out three variables, we set their polarities based on the clause distribution that are controlled by the parameters $ p_{1} $ and $ p_{2} $, which corresponds the code at Lines 9--16. With probability  $ p_{1} $, $ p_{2} $, and $ p_{3} = 1-p_{1}-p_{2} $, we generate a clause of Type 1, Type 2, and Type 3 respectively. In this procedure, we do not generate clauses of Type 0, which ensures the existence of the predefined solution $ s $.
	
	\section{Experiments}
	\label{sec-experiments}
	In this section, we first describe our experimental settings, including the selection of solvers that are used to evaluate the hardness of generated formulas, the generation of 3-SAT formulas, and the test platform (i.e., StarExec) on which the selected solvers are run. 
	Then, we graphically present and analyze the experiment results from different angles. 
	
	\subsection{Experimental settings}
	\subsubsection{The selection of solvers}
	In the top 10 solvers of the Random Satisfiable Track of the 2018 SAT Competition, one solver (Sparrow2Riss-2018, which ranked first) combines the SLS strategy and the CDCL strategy, four solvers (gluHack, \wrapletters{glucose-3.0\_PADC\_10\_NoDRUP}, \wrapletters{glucose-3.0\_PADC\_3\_NoDRUP}, and expGlucoseSilent, which came second, third, fourth, and fifth in turn) are primarily based on CDCL strategy, and five solvers (CPSparrow, dimetheus, probSAT, YalSAT, and lawa, which came sixth, seventh, eighth, ninth, and tenth) are primarily based on SLS strategy. 
	
	The same type of solvers have the similar behaviors on the same SAT formula, so we select two solver to evaluate the hardness of SAT formulas: one from the above CDCL solvers, and one from the above SLS solvers.
	Consequently, we select gluHack (came first in the CDCL solvers) and CPSparrow (came first in the SLS solvers), and thus we can verify different behaviors of currently top CDCL solvers and SLS solvers on SAT formulas generated by the proposed generating algorithm. 
	The SAT solver Sparrow2Riss-2018 (came first) is not selected, because it poses inconvenience of explaining its behavior for its combination of the SLS strategy and the CDCL strategy.
	
	We obtain the source code of gluHack and CPSparrow from the web site of the 2018 SAT Competition. 
	The parameter settings of these two SAT solvers have been tuned by the solver authors to obtain almost optimal performances in the 2018 SAT Competition. 
	Therefore, we adopt the same parameter settings in our experiments with those in the 2018 SAT Competition.
	The details are shown in \cite{heule2018proceedings} for gluHack and \cite{belov2014proceedings} for CPSparrow.
	
	\subsubsection{The generation of 3-SAT formulas}
	The parameter settings for the generation of 3-SAT formulas that are used in our experiments are shown as follows. 
	
	\begin{itemize}
		\item $ p $: $ 0.0-1.0 $ with the step size of $ 0.1 $. The default value is 0.3, which corresponds to a lower rate of intra-community clauses.
		\item $ \alpha $: $ 0.0-1.0 $ with the step size of $ 0.1 $. The default value is 1.0, which means there are not inter-community variables.
		\item $ \beta $ or ($ p_{1}, p_{2} $): The settings are shown in \myTable \ref{tab-settingp}. Note that the minimum of $ \beta $ is $ \frac{1}{3} $, where only one literal is true in each clause. The setting of ($ p_{1}, p_{2} $) corresponding to 0.5 of $ \beta $ (at this point, the numbers of true and false literals are equal) is called the balance setting. Below the balance setting, we set $ \beta $ to $ 0.35-0.50 $ with the step size of 0.05, and above that, we set $ \beta $ to $ 0.50-0.95 $ with the step size of 0.15. Then according to the setting of $ \beta $, we set the values of ($ p_{1}, p_{2} $). One setting of $ \beta $ corresponds to many pairs of ($ p_{1}, p_{2} $) (with a constraint that the sum of $ p_{1} $ and $ p_{2} $ must be no greater than 1). If ($ p_{1}, p_{2} $) is seen as a point, then these points constitute a line. Without loss of generality, in our experiments, ($ p_{1}, p_{2} $) is set to the midpoint of the line. The default value of $ \beta $ is 0.5, which is the balance setting.
		\begin{table}
			\renewcommand{\arraystretch}{1.3}
			\caption{The settings of ($ p_{1}, p_{2} $)}
			\label{tab-settingp}
			\centering
			\begin{tabular}{ccc}
				\toprule
				\bfseries ($ p_{1}, p_{2} $) &\bfseries $ \beta $ \\
				\midrule
				(0.9625, 0.0250) & 0.35 \\
				(0.8500, 0.1000) & 0.40 \\
				(0.7375, 0.1750) & 0.45 \\		
				(0.6250, 0.2500) & 0.50 \\
				(0.2875, 0.4750) & 0.65 \\
				(0.1500, 0.3000) & 0.80 \\
				(0.0375, 0.0750) & 0.95 \\	
				\bottomrule
			\end{tabular}
		\end{table}
		\item $ r $: $ 3.0-6.0 $ with the step size of 0.1. The default value is 4.5, which is around the phase transition point with regard to random uniform-3-SAT formulas.
		\item $ n $: $ 300-1600 $ with the step size of 50. The default value is 500.
		\item $ c $: $ 3-30 $ with the step size of 1. The default value is 20.
		\item $ s $: The predefined solution is randomly generated every time before generating a SAT formula.
	\end{itemize}
	
	Because of the randomness of our generating algorithm, in order to obtain a more accurate measurement of the hardness of SAT formulas that share the same group of parameter settings, we randomly generate $ 50 $ SAT formulas for every group of parameter settings.
	
	\subsubsection{The runtime platform}
	StarExec is a cross community logic solving service, that brings huge convenience to the experimental evaluation of SAT solvers. In our experiment, we first upload the source code of gluHack and CPSparrow to the StarExec. After the solvers are built on StarExec, we upload the files of SAT formulas generated by the proposed generating algorithm under different parameter settings. Then, we create jobs to run selected solvers on the generated SAT formulas. The parameter settings of jobs on StarExec are as follows.
	
	\begin{itemize}
		\item pre processor: none.
		\item post processor: checksat.
		\item woker queue: all.q(1).
		\item wallclock timeout: 1800 seconds. This is the maximum value allowed on StarExec.
		\item CPU timeout: 7200 seconds. This is also maximum value allowed on StarExec. 
		\item maximum memory: 24 GB. This setting is sufficient for runnings of our jobs; The evidence is that we do not get the state of ``memout", which means the running out of memory, in our experiments.
	\end{itemize}
	
	The CPU time represents the solving runtime of SAT formulas.
	Therefore, we use CPU time to represent the hardness of SAT formulas.
	More CPU time means higher hardness of SAT formulas.
	
	Although the CPU timeout is set to the allowed maximum value (i.e., 7200 seconds), the usage time of CPU would be less than the value of wallclock timeout (i.e., 1800 seconds) when ignoring the timing error. This is because gluHack and CPSparrow are serial programs, despite that they are run on quad-processors on StarExec. According to the analyses above, the upper-bound limit on the solving runtime for a SAT formula is the wallclock timeout (i.e., 1800 seconds). 
	
	\subsubsection{Processing experimental data}
	After jobs are completed, we obtain the experimental results from StarExec. For SAT formulas that are not successfully solved within the wallclock timeout, the corresponding CPU time cannot represent their solving hardness. However, considering that almost all values of the CPU time under case of wallclock timeout are very nearest to the value of wallclock timeout (the evidence is that the average CPU time of SAT formulas under cases of wallclock timeout is 1797 seconds, which could be easily calculated out from our experiment results), so we still use these CPU times to represent the solving runtimes.
	
	For the $ 50 $ SAT formulas that share the same group of parameter settings, we average the corresponding CPU times to obtain the hardness measurement.
	
	\subsection{Experimental results}
	
	In this section, we study the effects of various parameters in our algorithm, including $ p $, $ \alpha $, $ \beta $, $ r $, $ n $, and $ c $, on the hardness of SAT formulas generated by our generating algorithm with regard to gluHack and CPSparrow.
	We first schematically present our experimental results, then analyze the results.
	
	\subsubsection{The effect of $ p $ and $ \beta $}
\begin{figure}
		\centering
		\includegraphics[width=1\linewidth]{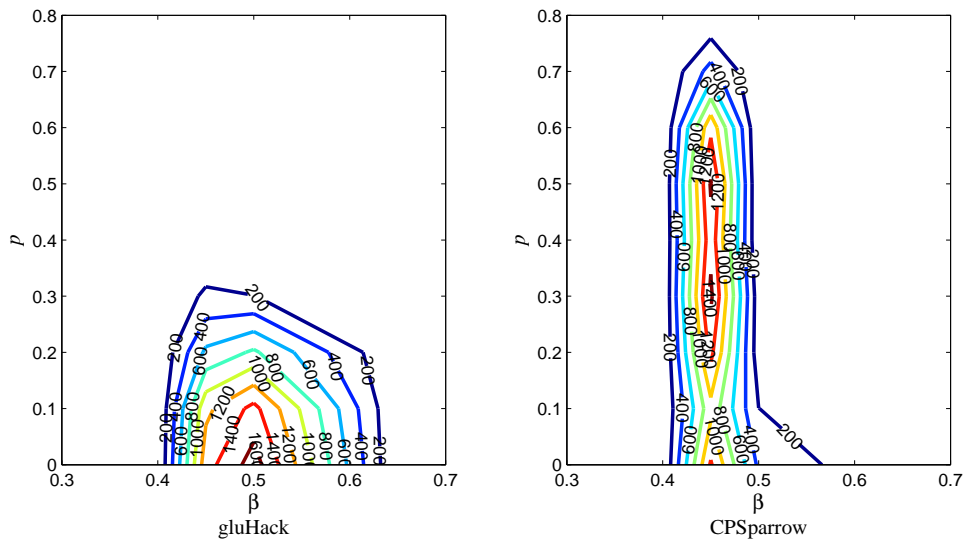}
		\caption{Contour plot of solving runtimes versus $ p $ and $ \beta $ when $ \alpha = 1 $, $ r = 4.5 $, $ n = 500 $, and $ c = 20 $}
		\label{fig:p}
	\end{figure}

	In this subsection, we study the effect of the parameters $ p $ and $ \beta $. 
	We fix parameters $ \alpha $, $ r $, $ n $, $ c $ to the default values 
	to observe how the solving runtimes change 
	under different combinations of $ \beta = [0.35, 0.40, 0.45, 0.50, 0.65, 0.80, 0.95] $ and $ p = [0.0, 0.1, 0.2, 0.3, 0.4, 0.5, $ $ 0.6, 0.7, 0.8, 0.9, 1.0] $. The contour plot of solving runtimes versus $ p $ and $ \beta $ is shown in \myFigure \ref{fig:p}.
	
	Our observations and analyses are as follows.
	\begin{itemize}
		\item[1)] Observations for gluHack: 
		For a fixed value of $ \beta $, the solving runtime of SAT formulas decreases as $ p $ increases. 
		This tendency indicates that the intra-community clauses make SAT formulas easy-to-solve, and gluHack exploit intra-community clauses to solve SAT formulas, which is consistent with the already existed conclusion in \cite{giraldez2016generating}.
		
		For all values of $ p $, the peaks of solving runtimes locate at around $ \beta = 0.5 $. This tendency indicates that only at the balanced setting, SAT formulas are hard-to-solve with regard to gluHack; and biased polarities of literals lead to easy-to-solve SAT formulas. The reason is that the bias introduces more solutions when the number of clauses is fixed, which make gluHack quickly find a solution. For lower $ \beta $ (i.e., 0.35, 0.40) and larger $ \beta $ (i.e., 0.65, 0.80, 0.95), the resulting SAT formulas are all easy-to-solve. This is because all these settings correspond to biased distribution of the polarities of literals. At this point, the effect of $ \beta $ has suppressed that of $ p $.
		
		\item[2)] Observations for CPSparrow:	
		For a fixed value of $ \beta $, when $ p \le 0.5 $, $ p $ has little or no effect on the hardness of SAT formulas. 
		But, when $ p > 0.5 $, the corresponding SAT formulas become easy-to-solve. This tendency indicates that more than half of intra-community clauses help CPSparrow solve SAT formulas.
		
		For all values of $ p $, the peaks of solving runtimes locate at around $ \beta = 0.45 $ instead of $ \beta = 0.50 $. The reason is that the mis-guidance caused by the biased literals \cite{liu2014p}. Note that according to the hardness level function of the $ p $-hidden algorithm \cite{liu2014p}, the 3-SAT formulas for lower $ \beta $ (i.e., 0.35) should be harder-to-solve with regard to SLS solvers, but in the right subplot of \myFigure \ref{fig:p}, they are not; this is because when $ r $ is lower than the required value (16.3 in current $ ({p}_{1}, {p}_{2}) $), many solutions except for the predefined solution are brought into the formula \cite{liu2014p}.
		However, when $ \beta $ is set to larger value, even if $ r $ is set to larger value, the resulting formulas are still easy to solve with regard to CPSparrow, because the mis-guidance disappears at this point.
		
		There are one exception in the right subplot of \myFigure \ref{fig:p}: valley at $ p = 0.1 $ and $ \beta = 0.45 $. This might be caused by the instability (the random selection of variables to flip) of CPSparrow.
		
		\item[3)] Comparisons: The intra-community clauses are exploited better by gluHack than CPSparrow. The values of $ \beta $ for hard-to-solve SAT formulas with regard to gluHack and CPSparrow are different (0.50 and 0.45 respectively), which indicates that biased polarities of literals have different effects on gluHack and CPSparrow.
		When at the balanced setting, CPSparrow is stronger than gluHack.

	\end{itemize}
	
	\subsubsection{The effect of $ p $ and $ \alpha $}
\begin{figure}
		\centering
		\includegraphics[width=1\linewidth]{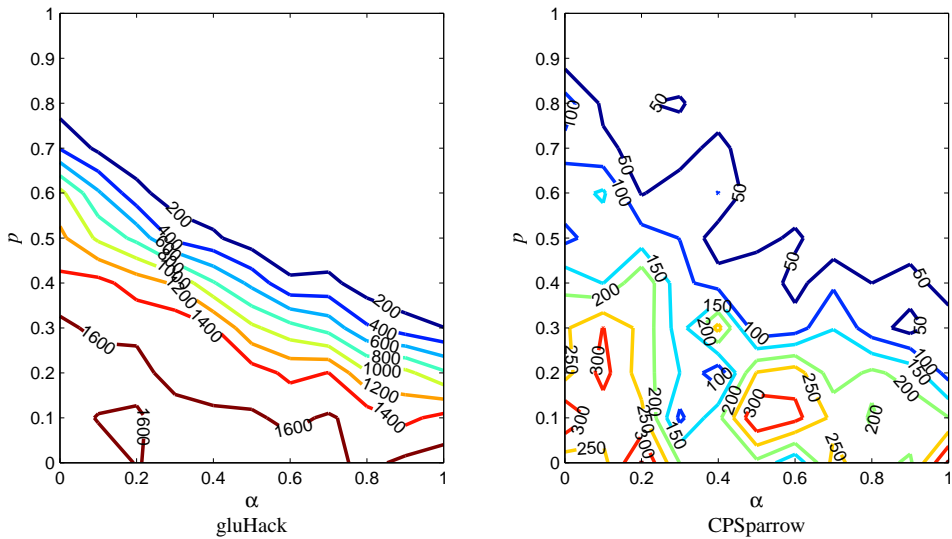}
		\caption{Contour plot of solving runtimes versus $ p $ and $ \alpha $ when $ \beta = 0.5 $, $ r = 4.5 $, $ n = 500 $, and $ c = 20 $}
		\label{fig:alpha}
	\end{figure}

	In this subsection, we study the effect of the parameters $ p $ and $ \alpha $.
	We fix parameters $ \beta $, $ r $, $ n $, $ c $ to the default values
	to observe how the solving runtimes change
	under different combinations of $ \alpha = [0.0, 0.1, 0.2, 0.3, 0.4, 0.5, $ $ 0.6, 0.7, 0.8, 0.9, 1.0] $ and $ p = [0.0, 0.1, 0.2, 0.3, 0.4, 0.5,$  $ 0.6, 0.7, 0.8, 0.9, 1.0] $.
	The contour plot of solving runtimes versus $ p $ and $ \alpha $ is shown in \myFigure \ref{fig:alpha}.
	
	Our observations and analyses are as follows.
	\begin{itemize}
		\item[1)] Observations for gluHack: 
		For a fixed value of $ p $, the solving runtime of SAT formulas decrease as $ \alpha $ becomes larger. This tendency indicates that the inter-community variables make SAT formulas hard-to-solve, and gluHack exploit intra-community variables to solve SAT formulas. When $ p > 0.8 $, lower $ \alpha $ does not lead to hard-to-solve SAT formulas, which is because the effect of $ p $ has suppressed that of $ \alpha $.
	
		\item[2)] Observations for CPSparrow: 
		With regard to CPSparrow, SAT formulas are all easy-to-solve (the maximum of solving runtimes is 357 seconds), and $ \alpha $ has little or no effect on the hardness of SAT formulas, which indicates that CPSparrow does not make use of the intra-community variables.
		
		\item[3)] Comparisons: The intra-community variables are exploited better by gluHack than CPSparrow. The intra-community variables may help gluHack find conflicts, so that it could find solutions quickly. 
	\end{itemize}
	
	\subsubsection{The effect of $ r $}
\begin{figure}
		\centering
		\includegraphics[width=0.9\linewidth]{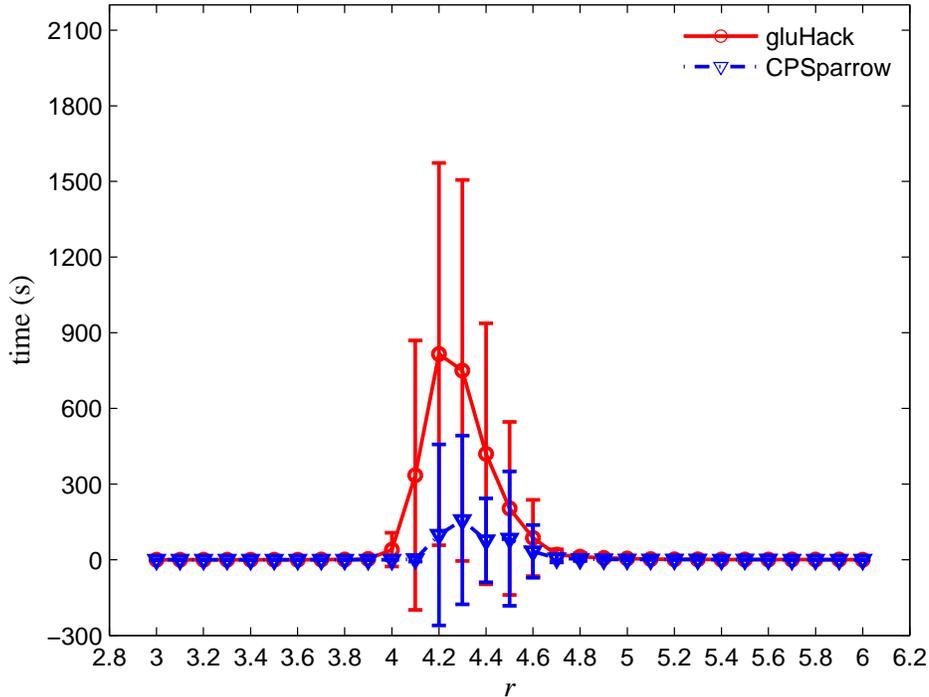}
		\caption{The effect of $ r $ when $ p  = 0.3 $, $ \alpha = 1.0 $, $ \beta = 0.5 $, $ n = 500 $, and $ c = 20 $}
		\label{fig:r}
	\end{figure}

	In this subsection, we study the effect of the parameter $ r $.
	We fix parameters $ p $, $ \alpha $, $ \beta $, $ n $, $ c $ to the default values
	to observe how the solving runtimes change as $ r $ changes,
	and draw the corresponding line plot of the solving runtimes versus $ r $.
	Also, the degree of dispersion of solving runtimes is drawn into the plot. The resulting plot is shown in \myFigure \ref{fig:r}.
	
	Our observations and analyses are as follows.
	\begin{itemize}
		\item[1)] When $ r $ locates around 4.25, 
		the SAT formulas are harder-to-solve with regard to both gluHack and CPSparrow. 
		The value of $ r $ is consistent with the phase transition point (estimated to be around 4.26) 
		for the random uniform-3-SAT formulas. 
		The random uniform-3-SAT formulas do not have a solution with high probability when $ r $ is above the phase transition point, 
		while the formulas generated by our generating algorithm always have at least one solution.
		\item[2)] Under the current parameter settings, 
		CPSparrow has stronger power than gluHack for solving these SAT formulas.
	\end{itemize} 
	
	\subsubsection{The effect of $ n $}
\begin{figure}
		\centering
		\includegraphics[width=0.9\linewidth]{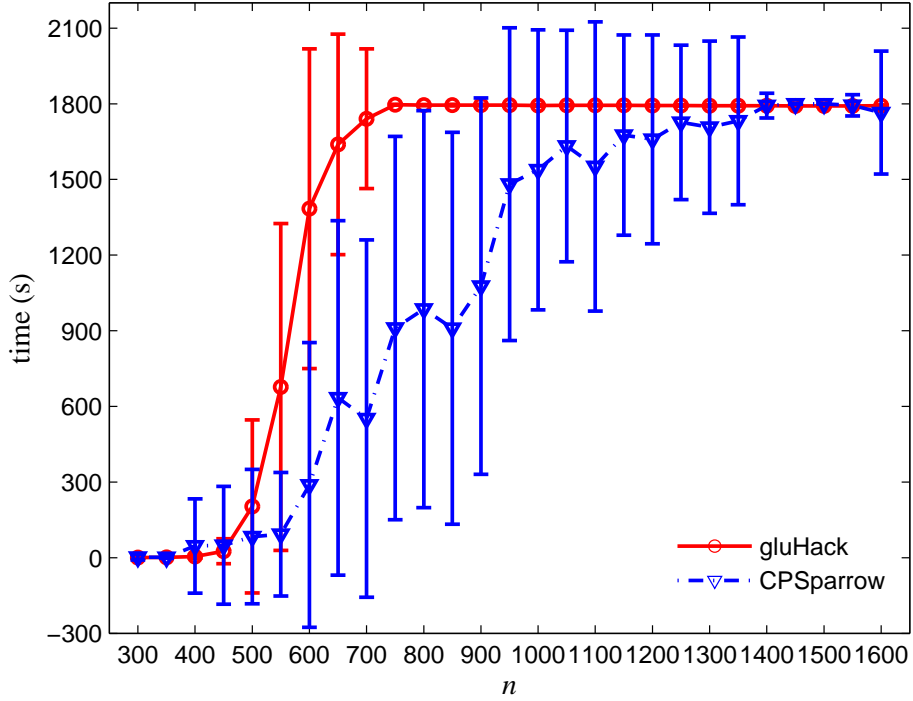}
		\caption{The effect of $ n $ when $ p  = 0.3 $, $ \alpha = 1.0 $, $ \beta = 0.5 $, $ r = 4.5 $, and $ c = 20 $}
		\label{fig:n}
	\end{figure}

	In this subsection, we study the effect of the parameter $ n $, 
	and try to find the critical point where SAT formulas are not successfully solved 
	under the current parameter settings with regard to both gluHack and CPSparrow.
	We fix parameters $ p $, $ \alpha $, $ \beta $, $ r $, $ c $ to the default values
	to observe how the solving runtimes change as $ n $ changes,
	and draw the corresponding line plot of the solving runtimes versus $ n $.
	Also, the degree of dispersion of solving runtimes is drawn into the plot. The resulting plot is shown in \myFigure \ref{fig:n}.
	
	Our observations and analyses are as follows.
	\begin{itemize}
		\item[1)] What is beyond doubt is that the solving runtime increases as $ n $ increases, because larger $ n $ means larger size of problem.
		\item[2)] Under the current parameter settings, the critical point from which the corresponding formulas are not successfully solved in the bounded time (i.e., 1800s) is around 750 for gluHack, while that is around 1450 for CPSparrow. Consequently, the critical point for gluHack is much less than that for CPSparrow, which indicates that CPSparrow is good at solving the SAT formulas under the current parameter settings.
		\item[3)] It is easily seen that the degree of dispersion of solving runtimes for gluHack is much less than that for CPSparrow.
	\end{itemize}
	
	\subsubsection{The effect of $ c $}
\begin{figure}
		\centering
		\includegraphics[width=0.9\linewidth]{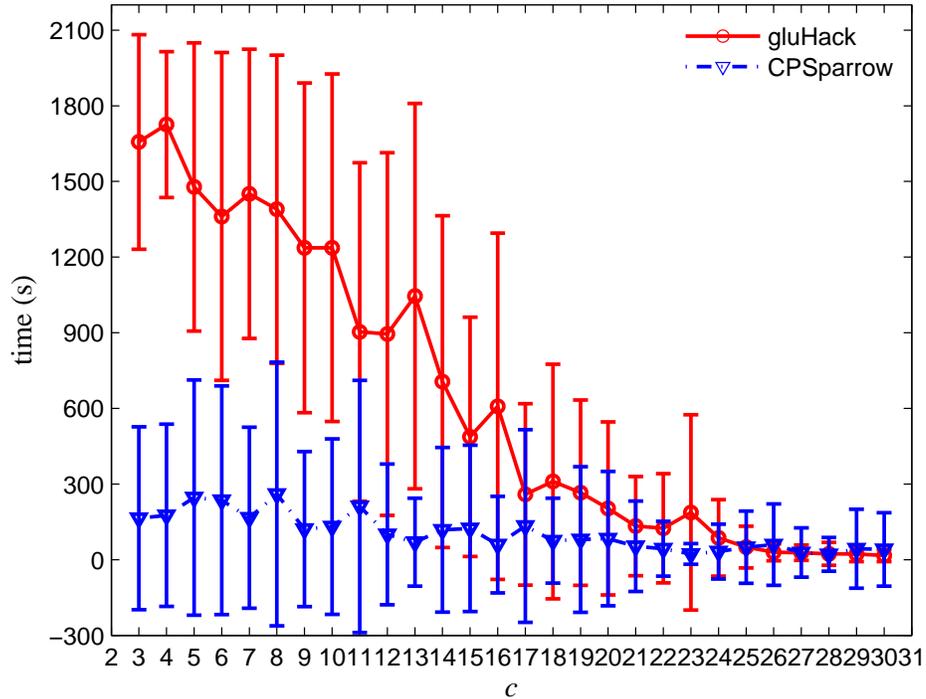}
		\caption{The effect of $ c $ when $ p  = 0.3 $, $ \alpha = 1.0 $, $ \beta = 0.5 $, $ r = 4.5 $, and $ n = 500 $}
		\label{fig:c}
	\end{figure}

	In this subsection, we study the effect of the parameter $ c $.
	We fix parameters $ p $, $ \alpha $, $ \beta $, $ r $, $ n $ to the default values
	to observe how the solving runtimes change as $ c $ changes,
	and draw the corresponding line plot of the solving runtimes versus $ c $.
	Also, the degree of dispersion of solving runtimes is drawn into the plot. The resulting plot is shown in \myFigure \ref{fig:c}.
	
	Our observations and analyses are as follows.
	\begin{itemize}
		\item[1)] With regard to $ gluHack $, as $ c $ increases, the hardness of SAT formulas decreases. This result further validates that the quality of community structures has a obvious effect on the hardness of SAT formulas with regard to gluHack.
		\item[2)] With regard to $ CPSparrow $, as $ c $ increases, the solving runtimes almost do not change.
			
	\end{itemize}
	
	\section{Discussions}
	\label{sec-discussions}
	
	SAT formulas generating algorithms with predefined solutions have many applications including information hiding \cite{liu2015hiding}, authentication \cite{dasgupta2008investigation,dasgupta2009biologically}, biometric recognition \cite{zhao2015negative}, and SAT-based cryptanalysis \cite{massacci2000logical,soos2009extending}. 
	
	The technique of negative databases \cite{esponda2004online,esponda2005negative,esponda2009negative} is strongly relevant to generating algorithms with predefined solutions, which converts a binary string to a group of binary strings, where the original string is seen as the predefined solution, and the each string in the resulting group of strings can be seen as a clause in a SAT formula. Negative databases protect information through preventing the group of binary strings being converted to the original string, which corresponds to solving a SAT formula, so generating hard-to-solve SAT formulas are extremely important for this technique.
	Since the randomness of our generating algorithm, the clauses in the resulting SAT formulas usually could not represent the whole complementary space of the predefined solution, so that the solution found by solvers might not be the predefined solution (i.e., $ s $, the input of our SAT formula generating algorithm). 
	However, finding the predefined solution is important in some applications, such as securely storing passwords through the technique of negative databases.
	There are some methods to avoid this problem. 
	For example, before generating a SAT formula, append the hash value of the predefined solution to the predefined solution.
	The hash value are calculated through a cryptographic hash function such as SHA-1 and SHA-256.
	Then generate a SAT formula corresponding to the extended solution \cite{esponda2008hiding}. 
	Thus, we could verify whether the found solution is the predefined solution: When a solver finds a solution from SAT formulas generated by our algorithm, check whether the values of the tail variables (the number of tail variables is dependent on the cryptographic hash function adopted above) in the found solution are the hash value of the front variables. 
	If success, the found solution is the predefined solution; otherwise, not.
	
	In our generating algorithm, only one predefined solution is considered. However, when replacing the $ p $-hidden algorithm in our generating algorithm with the $ m $-hidden algorithm \cite{liu2015hiding} or the extended $ K $-hidden algorithm (i.e., extend the $ K $-hidden algorithm \cite{zhao2015fine,zhao2017experimental} to generate SAT formulas with multiple solutions), the modified algorithm could generate SAT formulas with multiple solutions. Furthermore, the modified algorithm could be used to study the combined effect of community structures and multiple predefined solutions on the hardness of SAT formulas.
	
	\section{Conclusions and future work}
	\label{sec-conclusions}
	
	In this paper, we propose a generating algorithm of 3-SAT formulas with a predefined solution, which combines the features of community structures and clause distributions. We study the effect of the quality of community structures and clause distributions on the hardness of resulting formulas with regard to gluHack and CPSparrow through extensive experiments.
	
	In the future, we will study the reasonable construction approach of community structures (may be signed network \cite{shang2017multi,gomez2009analysis}) corresponding to SAT formulas that simultaneously considers variables and their polarities, so that we could study more natures of community structures corresponding to SAT formulas based on graphs with complete information of SAT formulas.
	
	\acks{
		This study is partially supported by the National Natural Science Foundation of China (No. 61175045). Wenjian Luo is the corresponding author. The source code of the proposed 3-SAT formula generating algorithm and the experimental results are available at:  \url{https://github.com/YaminHuPaperCode/Community-based-SAT-Formulas.git}.
	}
	
	\vskip 0.2in
	\bibliography{community}
	\bibliographystyle{theapa}
\end{document}